\documentclass[letterpaper, 10 pt, conference]{ieeeconf}  
\IEEEoverridecommandlockouts                              
\usepackage{graphicx} 
\usepackage{epsfig} 
\usepackage{mathptmx} 
\usepackage{times} 
\usepackage{amsmath} 
\usepackage{amssymb}  
\usepackage[colorlinks=true, allcolors=blue]{hyperref}
\usepackage{multirow}
\usepackage{gensymb}
\usepackage[utf8x]{inputenc}
\usepackage{float}
\usepackage{subfigure}

\title{DXSLAM: A Robust and Efficient Visual SLAM System with Deep Features}
\author{Dongjiang Li$^{1,2}$, Xuesong Shi$^{3*}$, Qiwei Long$^{1,2}$, Shenghui Liu$^3$, Wei Yang$^{2}$, Fangshi Wang$^{2}$, Qi Wei$^{1}$, Fei Qiao$^{1*}$
\thanks{$^{1}$Tsinghua University, Beijing, 100084 China.}
\thanks{$^{2}$Beijing Jiaotong University, Beijing, 100044 China.}
\thanks{$^3$Intel Corporation, China.}
\thanks{$^{*}$Corresponding authors: xuesong.shi@intel.com, qiaofei@tsinghua.edu.cn.}
}
\begin{document}
\maketitle

\begin{abstract}
A robust and efficient Simultaneous Localization and Mapping (SLAM) system is essential for robot autonomy. For visual SLAM algorithms, though the theoretical framework has been well established for most aspects, feature extraction and association is still empirically designed in most cases, and can be vulnerable in complex environments. This paper shows that feature extraction with deep convolutional neural networks (CNNs) can be seamlessly incorporated into a modern SLAM framework. The proposed SLAM system utilizes a state-of-the-art CNN to detect keypoints in each image frame, and to give not only keypoint descriptors, but also a global descriptor of the whole image. These local and global features are then used by different SLAM modules, resulting in much more robustness against environmental changes and viewpoint changes compared with using hand-crafted features. We also train a visual vocabulary of local features with a Bag of Words (BoW) method. Based on the local features, global features, and the vocabulary, a highly reliable loop closure detection method is built. Experimental results show that all the proposed modules significantly outperforms the baseline, and the full system achieves much lower trajectory errors and much higher correct rates on all evaluated data. Furthermore, by optimizing the CNN with Intel OpenVINO toolkit and utilizing the Fast BoW library, the system benefits greatly from the SIMD (single-instruction-multiple-data) techniques in modern CPUs. The full system can run in real-time without any GPU or other accelerators. The code is public at \url{https://github.com/ivipsourcecode/dxslam}.

\end{abstract}

\section{Introduction}
The problem of Simultaneous Localization and Mapping (SLAM) has seen great progress over the last decades \cite{cadena2016past}.
Among various visual SLAM algorithms, feature-based ones are prevalent in the robot industry for their efficiency and scalability in long-term robot deployments. However, most existing SLAM systems rely on hand-crafted visual features such as SIFT \cite{LoweDistinctive}, Shi-Tomasi \cite{shi1994good} and ORB \cite{rublee2011orb}, which may fail to provide consistent feature detection and association results in complex environments. For example, ORB-SLAM2 \cite{mur2017orb} frequently fails to recognize previously visited scenes when either the scene or the viewpoint has been changed \cite{shi2019we}.

On the other hand, for most other computer vision tasks, feature extraction based on deep convolutional neural networks (CNNs) have replaced the hand-crafted features to dominant related research and applications. Trained with large amount of diversified data, CNNs are able to learn feature representations robust against changes in illumination, background and viewpoint. Though most works with CNNs deal with image region-level features (e.g. semantics), there are works focusing on learning pixel-wise features \cite{detone2018superpoint}\cite{dusmanu2019d2}\cite{tang2019gcnv2}, usually referred to as \textit{local features} or \textit{local descriptors}. Though it has been shown that those deep features are superior to hand-crafted ones in many aspects, there have not been many works utilizing them in visual SLAM systems. Besides the required efforts on system tuning, one reason may be that those deep CNNs require a GPU or other hardware accelerators for real-time inference, which may be impractical for robots or other mobile systems.

For some of the closely related topics of SLAM, including visual (re-)localization and loop closure detection (LCD), recent works have a much stronger preference to deep CNN-based approaches, as the performance gap from conventional methods is significant. Research in those fields usually requires extracting \textit{global features} of each image. This can be done by aggregating local deep features with e.g. the bag-of-words (BoW) method \cite{yue2019robust}, or with an end-to-end CNN inference, e.g. NetVLAD \cite{arandjelovic2016netvlad}.

\begin{figure*}[t!]
\centering
\includegraphics[scale=0.6]{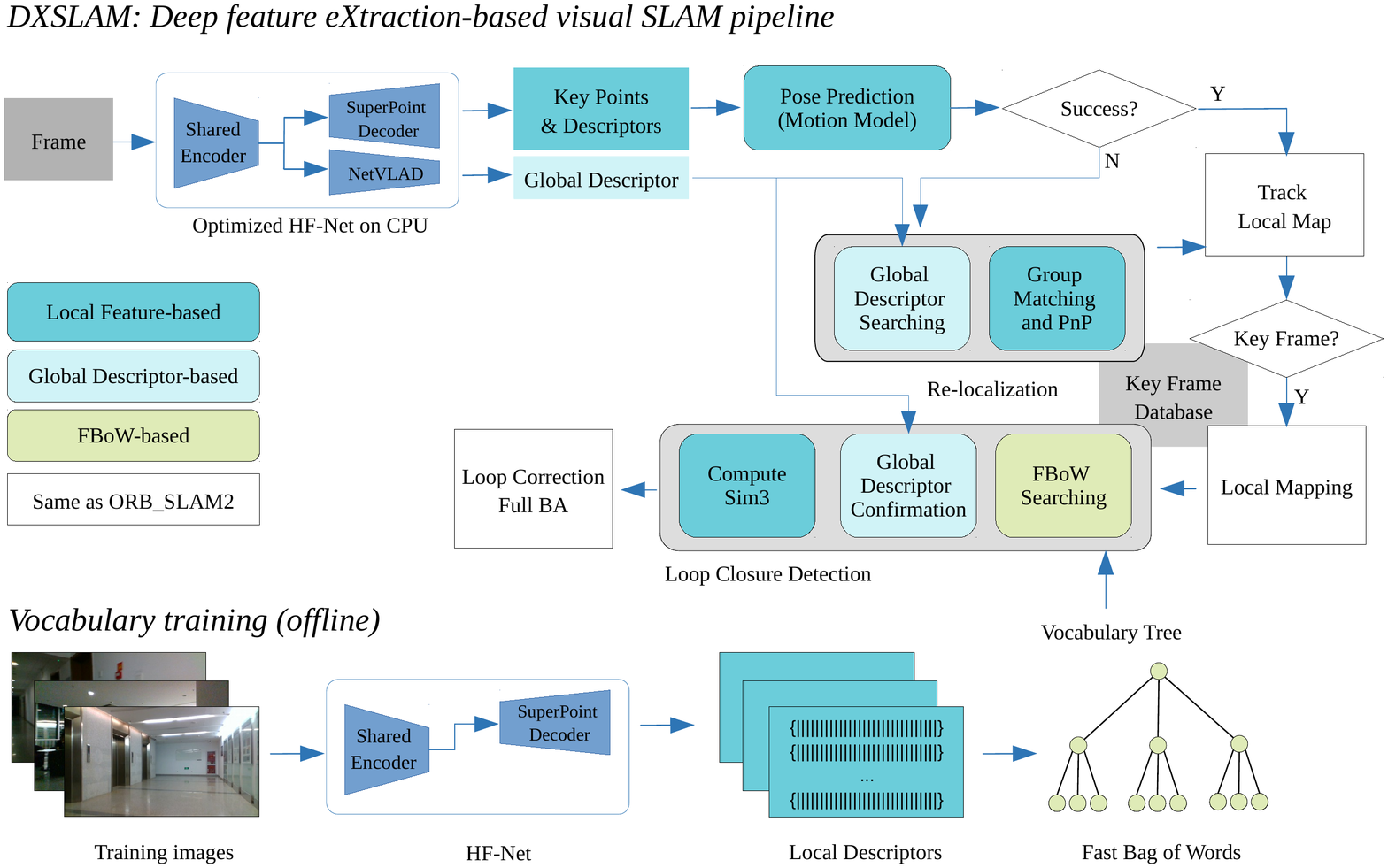}
\vspace{-2cm}
\caption{The framework of the proposed visual SLAM system. The pipeline is largely the same with ORB-SLAM2 \cite{mur2017orb}, with deep features incorporated into various modules of the system.}
\label{fig:system-framework}
\end{figure*}

In this paper, we present a novel visual SLAM system based on learned features aiming to enhance the lifelong localization capability in changing environment. The system, named DXSLAM, uses a Deep CNN to eXtract both local features and global features from each image frame. Those features are then fed into a modern SLAM pipeline for pose tracking, local mapping, LCD and re-localization, as shown in Fig. \ref{fig:system-framework}. The contributions of this paper include:

\begin{itemize}
\item It presents a full SLAM system with loop closure, global optimization and re-localization, all based on features from a state-of-the-art deep CNN, providing much more robustness against environmental and viewpoint changes than SLAM with hand-crafted features.
\item A robust re-localization method with global feature-based image retrieval and group matching is proposed. It achieves a much higher success rate than conventional BoW-based methods, with a much lower computation cost.
\item A reliable LCD method based on both global and local features is proposed. A novel visual vocabulary is trained to aggregate local features.
\item The proposed SLAM system is optimized for modern CPUs by using Intel OpenVINO toolkit for feature extraction and using Fast BoW \cite{munoz2020ucoslam} in LCD. To our knowledge, it is the first deep feature-based SLAM system that can run in real-time without GPUs.
\end{itemize}

\section{Related Works}
In this section, we review existing works that are mostly related to our contributions. We refer readers interested in the general aspects of SLAM algorithms to \cite{cadena2016past} for a comprehensive overview, and \cite{huang2019survey} for a complete recent survey.

\subsection{Feature Extraction for Visual SLAM}
Many visual SLAM pipelines start from detecting keypoints from an image frame and matching them with those from a previous keyframe or from a map by the similarity of their descriptors. Among the various keypoint features in computer vision, Shi-Tomasi \cite{shi1994good} and ORB \cite{rublee2011orb} are the mostly used ones by visual SLAM algorithms (e.g. MonoSLAM \cite{davison2007monoslam}, ORB-SLAM2 \cite{mur2015orb}, VINS-Mono \cite{qin2018vins}) for their balanced effectiveness and efficiency.

Not long after the begin of the deep learning era, it was found that features from the last layer of a deep CNN trained with ImageNet can outperform hand-crafted local features, even though the CNN is not trained for such local representations \cite{fischer2014descriptor}. Since then, works on training CNNs specifically for local feature extraction have been presented \cite{han2015matchnet}\cite{balntas2016learning}\cite{zagoruyko2015learning}\cite{mishchuk2017working}. These CNNs take a local image patch as input and outputs a descriptor of this patch. Though their performance against pre-trained deep CNNs may be uncertain, both approaches consistently outperform hand-crafted features \cite{dai2019comparison}.

For deep learning, keypoint detection is more difficult than patch description, in the sense that the notation of keypoint is semantically ill-posed, making direct data annotation infeasible. DeTone et al. addresses this problem by proposing a self-supervised learning approach and using it to train a fully-convolutional neural network for joint keypoint detection and description (SuperPoint) \cite{detone2018superpoint}. Another network GCN is trained for the same capabilities, but with a supervison of visual odometry \cite{tang2018geometric}. Dusmanu et al. designed D2-Net in which the knowledge of keypoint detection and description can be further shared \cite{dusmanu2019d2}.

Our SLAM system is built upon above works. In DXSLAM, we empirically selected HF-Net \cite{sarlin2019coarse} to provide SuperPoint-like keypoint detection and description, though other CNN-based approaches would also work in the proposed system. A most similar work is GCNv2 \cite{tang2019gcnv2}, which also incorporates deep local features into the pipeline of ORB-SLAM2.

It is worth noting that there are CNNs for end-to-end pose estimation \cite{wang2017deepvo} \cite{kendall2015posenet} \cite{melekhov2017relative} \cite{detone2016deep}, making feature extraction an implicit task hidden in the networks. Though their accuracy is not yet competitive, it is interesting to further investigate on such approaches as they have common traits with biological localization.

\subsection{Loop Closure Detection}
LCD requires to recognize a previous visited place from current camera measurements \cite{angeli2008fast}. The mostly used LCD method for real-time SLAM is to train a visual vocabulary with a tree structure based on local features to improve the image retrieval efficiency \cite{nister2006scalable}, which can aggregate local features and eventually give a vector representation of an image. One of the most successful applications of this method is FAB-MAP, which employs SURF and BoW to train vocabulary for place recognition and demonstrates robust performance against viewpoint changes \cite{cummins2008fab}. FBoW (Fast Bag-of-Words) is an optimized implementation of BoW \cite{munoz2020ucoslam}. It uses the single-instruction-multiple-data (SIMD) techniques on x86 CPUs, achieving significant speed-up of vocabulary loading and matching.

In order to utilize the representation ability of CNNs, recent works attempt to combine them with BoW-based LCD algorithms \cite{xia2017evaluation}. \cite{yue2019robust} proposed a robust LCD algorithm based on bag of SuperPoints and graph verification. However, its robustness tends to decrease under low illumination condition in our experiments. \cite{hou2018bocnf} presents the BoCNF method, which also uses CNN-based features to build a visual vocabulary and to detect loops in large-scale challenging environments. 

\subsection{Re-localization}
Re-localization is often formed as a pipeline of image retrieval followed by relative pose estimation, similar to LCD, but often with a much larger database of candidate images, and with more emphasis on high recall as opposed to high precision of LCD. For image retrieval, either local features with an aggregation method such as BoW or deep CNN-based global features can be used. The difference is that the aggregation policy in the latter is learned end-to-end from data, thus they generally can have better performance. The state-of-the-art method for aggregating local features within a CNN is using a NetVLAD layer \cite{arandjelovic2016netvlad}, which is modified from VLAD to be differentiable. In \cite{sarlin2019coarse}, HF-Net is proposed to bind a SuperPoint-like network with a NetVLAD network, with shared encoder layers to reduce computation cost and facilitate training. For pose estimation between the retrieved image and the query image, the PnP method \cite{kneip2011novel} within a RANSAC scheme \cite{fischler1981random} has been prevalent in both visual localization works \cite{sarlin2019coarse} and SLAM systems \cite{mur2017orb}.

\section{SLAM System with Deep Features}

\subsection{Overview}
The proposed SLAM system is shown in Fig. \ref{fig:system-framework}. The framework is similar to ORB-SLAM2 \cite{mur2017orb}. Differences originate from feature extraction, for which we use HF-Net \cite{sarlin2019coarse} to give both local features (key points and their descriptors) and global features (image descriptors) with a single CNN model. The local features are then incorporated into the localization and mapping pipeline. Based on the global features, an efficient re-localization module is built to rapidly re-localize at system initialization or tracking failures. To mitigate accumulated localization errors and achieve globally consistent mapping, a robust loop closure method is proposed. The method considers both the global features from HF-Net and local feature matching with a pre-trained BoW vocabulary.

To further improve the system efficiency, we adopt the FBoW method to get binary visual vocabulary, which significantly reduces system initialization time and improves matching efficiency compared with conventional BoW methods. We also optimized the HF-Net model and re-implement the inference process with Intel OpenVINO toolkit, to enable real-time feature extraction on CPUs.

\subsection{Feature Extraction}
We use a deep CNN, HF-Net \cite{sarlin2019coarse} to extract features from each image frame. In HF-Net, an image firstly passes through a shared encoder, and then goes into three parallel decoders predicting key point detection scores, dense local descriptors and a global image-wide descriptor, respectively. The first two decoders have the same architecture as SuperPoint \cite{detone2018superpoint}, and the global descriptor is computed by a NetVLAD layer \cite{arandjelovic2016netvlad}. This design enables HF-Net to give both local and global features with a single inference model, benefiting not only subsequent pose tracking, but also LCD and re-localization modules as described in the following subsections.

This design choice is not only driven by functionalities, but also experimental results showing that the features from HF-Net are superior than those from alternative deep CNN feature extractors for the SLAM system. Some of the results are shown in Section IV.

The original implementation of HF-Net is with TensorFlow. We optimize the pre-trained model with Model Optimizer from Intel OpenVINO toolkit, and re-implement model prediction with Inference Engine, which utilizes the SIMD operations on x86 CPUs. Most of the layers in HF-Net can be directly processed by the Model Optimizer, except the bilinear interpolation operation for local descriptor up-sampling, which is not yet supported, and thus we move it out of the network into a post-processing stage after OpenVINO inference.

\subsection{Vocabulary Training}
In the traditional BoW, K-D tree is adopted to accelerate the search process. However, if levels and nodes of the K-D tree are not assigned properly during the vocabulary training process, the vocabulary will not distinguish the features well. In order to solve this problem, we fully consider the relations between consequential training images and adopt an incremental manner to train vocabulary. During the training phase, keypoints and local descriptors are first extracted from consecutive image sequence by HF-Net. For each pair of adjacent images, we use the brute-force approach to match images. The matched local descriptors are supposed to belong to the same existing leaf node, namely visual word, and the unmatched features are assigned as new leaf node. Theoretically, we should place all matched descriptors under the mentioned existing word. However, considering the reliability of extracted features, we select the top 300 matched descriptors by corresponding key point detection scores. After all the training images are processed, we can get a series visual words and then cluster these words into parent nodes. In this paper, the OpenLORIS-Scene datasets \cite{shi2019we} are used to train the vocabulary. The trained vocabulary can properly quantify features into visual words and images finally are represented and matched using corresponding vectors depicting the histogram of words.

To improve the efficiency of feature matching and the overall SLAM system, we adopt the FBoW framework to build the whole vocabulary. The trained vocabulary with FBoW is in binary form, which is much more efficient to be loaded and used in feature matching. In our tests, the system initialization time of DXSLAM is about 40 milliseconds, while that of ORB-SLAM2 is about 6 seconds.

\subsection{Re-localization with Global Features}
ORB-SLAM2 employs a two-phase pipeline for re-localization: first retrieve candidate frames similar to the current frame with BoW feature matching, then a frame-to-frame method is adopted to estimate current frame pose by matching its local features and each candidate's, until all candidates have been traversed or the pose has been estimated. Failures can be caused by two problems:
\begin{itemize}
    \item The BoW method fails to retrieve any candidate frames.
    \item Not enough matched local features for pose estimation in the second phase.
\end{itemize}

We address the first problem by implementing a coarse image retrieval based on the learned global descriptors, which have been widely verified to be much more robust to environmental and viewpoint changes than the BoW method. The retrieved candidate frames are then used for group matching. By matching keypoints of the current frame to all the keypoints in group, the second problem can be greatly mitigated. Finally a standard RANSAC and PnP process is executed for each group with enough matched keypoints. The number of groups is usually two or three in our experiments.

\subsection{Loop Closure Detection with Multi-level Features}
LCD is important for the SLAM system to correct accumulated errors and build a consistent map. Though it can also be formed as an image retrieval problem, its requirement has an emphasis on precision rather than recall in re-localization, as a false loop closure may damage the map. Therefore, more strict criteria should be applied for LCD. In DXSLAM, we use both local and global descriptors to detect loops. For each new keyframe, local descriptors are first quantized into words by matching with the nodes of the pre-trained vocabulary tree and then the keyframe is represented by a visual vector. We select top $K$ of keyframes from database according to similarity score by calculating distance of visual vectors defined in \cite{nister2006scalable}. The similarity score between two frames with visual vectors $v_1$ and $v_2$ is defined as
\begin{equation}
    s(v_1,v_2)=\sum_{i=1}^N |v_{1,i}|+|v_{2,i}|-|v_{1,i}-v_{2,i}|.
\end{equation}

Because the BoW matching method aggregates local features by their distributions and discards their spatial relations, false matches can occur. In our system, metrics based on global descriptors can be a complementary criterion to address this issue. In the second phase of our LCD, a distance between the current frame and each of the above top $K$ candidates is calculated based on the inner product of their global descriptors. The candidate with the smallest distance, if below a predefined threshold, is reported as a detected loop.

\section{Evaluation}

In this section, we present evaluation results of each proposed module and the full SLAM system. We first perform a performance comparison between visual SLAM systems with different feature extractors. We then evaluate the proposed re-localization and loop detection methods on controlled data with separate challenge factors. Finally we evaluate the performance of the full DXSLAM system with lifelong SLAM datasets.

\begin{figure*}[t!]
\centering
\includegraphics[scale=0.64]{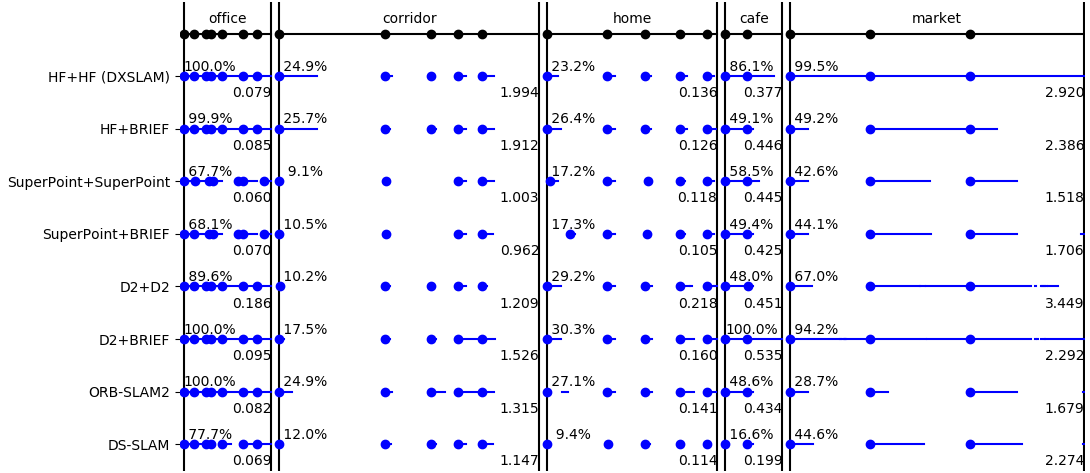}
\caption{Per-sequence testing results of visual SLAM (without re-localization and loop closure) with different features on the OpenLORIS-Scene datasets. An algorithm with name A+B means using A for keypoint detection and B for keypoint description. Each black dot on the top line represents the start of one data sequence. For each algorithm, blue dots and lines indicate successful initialization and tracking. The percentage value on the top left of each scene is the average correct rate, larger meaning more robust. The float value on the bottom right is the average ATE RMSE, smaller meaning more accurate.}
\label{fig:per-sequence-t-5}
\end{figure*}

\subsection{Datasets} 
The evaluation uses two types of data: SLAM evaluation datasets and LCD evaluation datasets.

\subsubsection*{TUM RGB-D \cite{sturm2012benchmark} and OpenLORIS-Scene \cite{shi2019we}} 
TUM RGB-D is the mostly used SLAM benchmark in literature. It provides a variety of data sequences with precise ground-truth trajectories. OpenLORIS-Scene is a recently published dataset providing real-world robotic data with more challenging factors including blur, featureless images, dim lighting, and significant environmental changes. The last can be a major challenge for re-localization. In this section, we mainly use OpenLORIS-Scene to evaluate algorithm robustness, while use TUM RGB-D to evaluate localization accuracy.

\subsubsection*{City Center and New College \cite{cummins2008fab}}
These two datasets are widely used for LCD evaluation. They are both collected with a mobile robot. The former includes many dynamic objects like pedestrians and vehicles. Besides, the sunlight, wind and viewpoint changes can pose challenges to LCD algorithms. The New College dataset includes not only dynamic objects, but also repeated visual elements, such as similar walls and bushes, which tend to cause false loop detections.

\subsection{Feature Evaluation}
In order to test the solidified HF-Net performance, we compare it with hand-crafted features and other state-of-the-art feature extraction methods based on deep CNNs, including SuperPoint \cite{detone2018superpoint} and D2-Net \cite{dusmanu2019d2}. We use the public available model weights released by the original authors for all three CNN models without any re-training or fine-tuning. We substitute feature extraction module of ORB-SLAM2 with each of the methods, and compare the localization performance on the OpenLORIS-Scene datasets (in a per-sequence evaluation fashion). Re-localization and LCD modules are disabled in this test. To have a deeper understanding of the each method's merits and shortcomings, we also explore different combinations of keypoints and descriptors, such as HF-Net keypoints with BRIEF discriptors \cite{calonder2010brief} (denoted as HF+BRIEF). The results are shown in Fig. \ref{fig:per-sequence-t-5}. It also includes the results of other feature-based SLAM systems from \cite{shi2019we} as baselines. All the algorithms in Fig. \ref{fig:per-sequence-t-5} are well comparable as they share the same framework of ORB-SLAM2, only differing in what features they use. DS-SLAM introduces a feature selection scheme to filter out the ORB keypoints on dynamic objects \cite{ds-slam}.

There are five scenes in the OpenLORIS datasets. The scenes of corridor and home contain totally featureless images (white walls), making it impossible to track over the whole trajectory with any visual feature only SLAM. Thus we focus on the results with the other scenes (office, cafe and home). It is obvious that the robustness of CNN-based visual SLAM is superior than ORB-SLAM2 and DS-SLAM. However, the performances of CNN-based methods are different from each other. 
For example, the combination of D2+BRIEF can improve the performance largely compared to D2+D2. However, this is not truth with the HF+HF. Compared to BRIEF, the local descriptor extracted from HF-Net can represent features more properly. SuperPoint tends to extract less features under low light illumination conditions like office scene which usually lead to track failure, but HF-Net does not encounter such problem. By various comparative experiments, it is verified that we adopted HF-Net as the front-end of proposed visual SLAM system is reasonable, which is beneficial to improve the system robustness under changed scenes.

\subsection{Re-localization Evaluation}
The office scene in the OpenLORIS-Scene datasets provides a set of controlled challenging factors to evaluate re-localization robustness against each factor. Testing results are shown in Table \ref{tab:reloc}, reporting re-localization scores as defined in \cite{shi2019we}. The proposed method outperforms ORB-SLAM2 on all cases. It fails to re-localize only when where is a significant viewpoint change (the robot in office-2 is moving total-oppositely against the trajectory of office-1), suggesting that scene recognition between largely different viewpoints should be an open problem. It is worth noting that the robustness against illumination challenges has been greatly improved.

\begin{figure}[t!]
\centering
\includegraphics[scale=0.14]{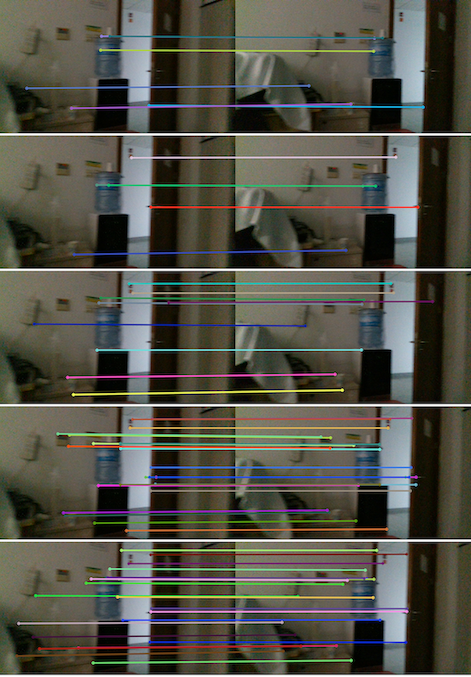}
\caption{Matching between a query frame (left) and a group of retrieved frames (right). Re-localization candidates are retrieved by global descriptor matching and then are used for group matching. The five images at the right side are retrieved candidates.}
\label{fig:relocalization}
\end{figure}

Fig. \ref{fig:relocalization} shows an example of the proposed re-localization method using group matching of retrieved keyframes to match features with current frame. In Fig. \ref{fig:relocalization}, the right of five images represent a group data and the left image is the current frame. We can find that each keyframe in the group uses its best quality key points to match. by this way, the quantity and quality of matched key points are both better than the frame-to-frame method.

\begin{table}[]
    \centering
   \caption{Re-localization Scores under Different Conditions from OpenLORIS-Scene Datasets} 
    \begin{tabular}{c c c c c c}
   \hline
    \textbf{Data: office-}& \textbf{1,2}& \textbf{2,4}& \textbf{2,5} & \textbf{1,6} & \textbf{2,7} \\
    \hline
    \textbf{Key factor}& \textbf{viewpt.}& \textbf{illum.}& \textbf{low light} & \textbf{objects} & \textbf{people} \\
    \hline
   ORB-SLAM2 & 0 & 0.764 & 0 & 0.716 & 0.997 \\
   DS-SLAM & 0 & 0 & 0 & 0.994 & 0.996 \\
   DXSLAM & 0 & 0.862 & 0.994 & 0.999 & 0.999 \\
   \hline
   \end{tabular}
   \label{tab:reloc}
\end{table} 

\subsection{Loop Closure Detection Evaluation}
\begin{figure*}[t!] 
	\centering  
	\subfigtopskip=2pt
	\subfigbottomskip=2pt 
	\subfigcapskip=-3pt
	\subfigure[New College dataset]{
		\label{level.sub.1}
		\includegraphics[height=5cm]{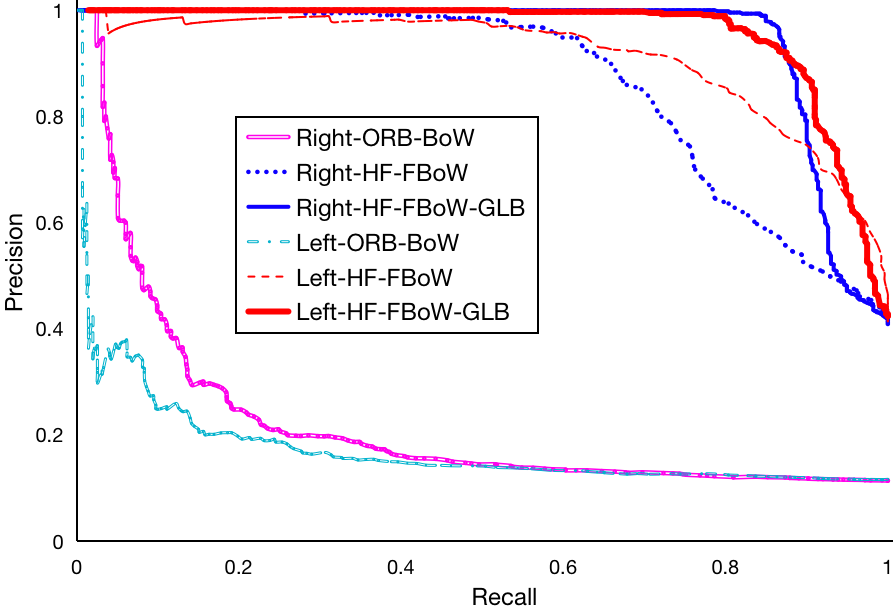}}
	\quad
	\subfigure[City Center dataset]{
		\label{level.sub.2}
		\includegraphics[height=5cm]{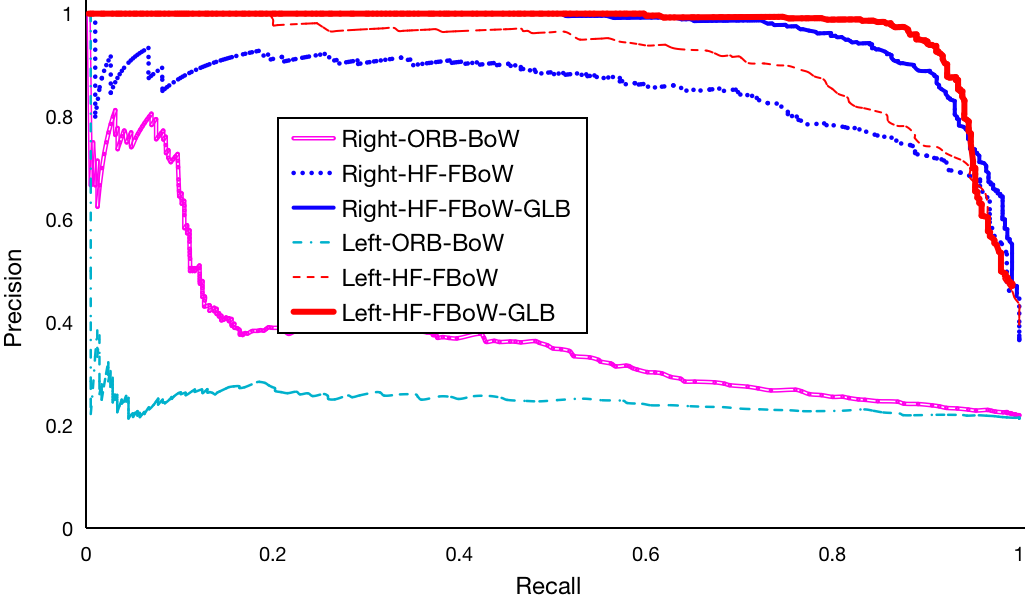}}
	\caption{The PR-Curves of LCD from the proposed method (HF-FBoW-GLB), the proposed method without using global features (HF-FBoW), and ORB-SLAM2}
	\label{fig:lcd}
\end{figure*}

The proposed LCD algorithm contains two parts as mentioned in section III E. It first retrieves loop candidates by trained vocabulary based on FBoW framework and then uses the spatial information provided by global descriptors to remove wrong matches. During the evaluation, our algorithm divides into two parts. One is the full LCD algorithm, denoted as HF-FBoW-GLB, and the other is denoted as HF-FBoW which selects the top 1 candidate by the similarity score. In order to compare the performance with the method in OBR-SLAM2, we also evaluate BoW with ORB features, denoted as ORB-BoW. The City Center and New College datasets contain two sets of images collected by a robot with left and right cameras, respectively. We test these three methods on both sets of camera data. The precision and recall results are shown in Fig. \ref{fig:lcd}. It is clear that the proposed method greatly improves over ORB-SLAM2, and that the combination of both local and global features also contribute a notable improvement.

\begin{figure*}[t!]
\centering
\includegraphics[scale=0.64]{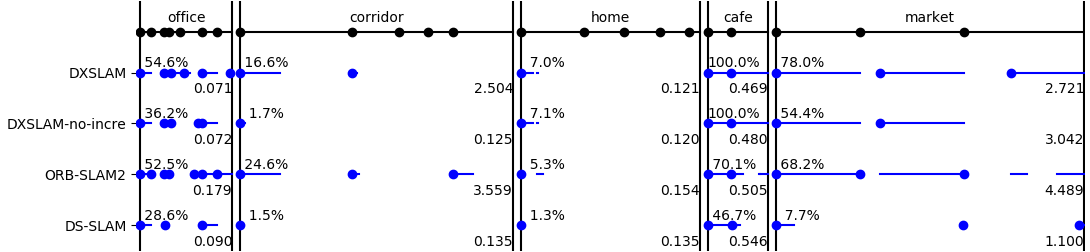}
\caption{Lifelong SLAM testing results with the OpenLORIS-Scene datasets. For each algorithm, blue dots and lines indicate successful initialization/re-localization and tracking. The percentage value on the top left of each scene is theaverage correct rate, larger meaning more robust. The float value on the bottom right is the average ATE RMSE, smaller meaning more accurate.}
\label{fig:lifelong-t-3}
\end{figure*}

\subsection{Full System Evaluation}

We evaluate the performance of DXSLAM on OpenLORIS-Scene datasets and TUM RGB-D datasets. It is tested in a lifelong SLAM fashion on the former, in which data sequences of the same scene are sequentially fed to the algorithm. The results are shown in Fig. \ref{fig:lifelong-t-3}. Besides the baselines of ORB-SLAM2 and DS-SLAM. We also trained another version of DXSLAM visual vocabulary without the incremental method, denoted as DXSLAM-no-incre. 

We continue to focus on the scenes of office, cafe and market. 
Compared to DXSLAM-no-incre, DXSLAM has more number of correct pose estimation and re-localization, which can evaluate that our built visual-vocabulary by incremental method is more efficient and reasonable. DXSLAM also achieves significant robustness toward changing environment. For example, compared to its original framework ORB-SLAM2, DXSLAM tracks more trajectory and ATE RMSE is far smaller.

\begin{table}[]
    \centering
    \caption{ATE RMSE Results on TUM RGB-D} 
    \begin{tabular}{c c c c}
    \hline
    Sequence & ORB-SLAM2 & DS-SLAM & DXSLAM \\ 
    \hline
    fr3\_walking\_xyz & 0.7521m & 0.0247m & 0.3088m \\
    fr3\_walking\_static & 0.3900m & 0.0081m & 0.0167m \\
    fr3\_walking\_half & 0.4863m & 0.0303m & 0.0759m\\
    fr3\_sitting\_static & 0.0087m & 0.0065m & 0.0068m\\
    fr3\_sitting\_half & 0.0208m & 0.0150m & 0.0153m\\
    \hline
    \end{tabular}
    \label{tab:tum}
\end{table} 

The accuracy of localization is also evaluated with TUM RGB-D. The results are shown in TABLE \ref{tab:tum}.
It can be seen that DXSLAM has a strong capability to resist dynamic environments, giving comparable results with DS-SLAM, though that dynamic features are not explicitly addressed in DXSLAM.

\subsection{Runtime Performance}

A high-power discrete GPU is usually preferred to execute deep CNN models, but they are unavailable on most robot platforms due to power and space constraint. We tested the performance of different feature extraction models both on a GPU machine and on a GPU-free mini-PC which are commonly used to build robot prototypes. The former is an Alienware laptop equipped with an NVIDIA GeForce GTX 1070 GPU (GPU TDP 150 Watts). The latter is an Intel NUC with an ultra-low power CPU Intel Core i7-10710U (CPU TDP 15 Watts, full system TDP 25 Watts). The results are shown in TABLE \ref{tab:runtime}. The data are the average time cost to process each image in the office1-1 sequence in the OpenLORIS-Scene datasets, all resized to 640x480 pixels. For HF-Net, we set the iterations and radius of the non-maximum suppression (NMS) both to be 1 for balanced efficiency and accuracy. All other models are with their default configuration. The results indicate that HF-Net is much more efficient than the other CNN models, as it consumes less compute time while giving more results (global descriptors). With OpenVINO optimization, the time cost of HF-Net on CPU is further reduced by 68\% to 46.2 ms (21.6 FPS), able to be incorporated into real-time SLAM systems. With our ROS implementation of DXSLAM, where feature extraction and SLAM run as two ROS nodes on the same machine taking live RGB-D stream as input, the final pose estimates are published at around 15 Hz on the Intel NUC. We expect that the performance can be further enhanced by e.g. CNN pruning, quantization and software optimization.

\begin{table}[]
\centering
\caption{Average Time of Feature Extraction (millisec per image)}
\begin{tabular}{c|c|c}
    \hline
         & CPU only & With GPU \\
    Method  & Core i7-10710U & GeForce GTX 1070 \\
          & (15W)          & (150W) \\
    \hline
    ORB         & 14.1      & - \\
    SuperPoint  & 387.5     & 28.8 \\
    D2-Net      & 2484.6    & 217.0 \\
    HF-Net      & 144.2     & 19.7 \\
    HF-Net with OpenVINO & 46.2 & - \\
    \hline
\end{tabular}
\label{tab:runtime}
\end{table}

\section{Conclusions}

In this paper, we show that deep CNN-based features can be well incorporated into modern SLAM systems, and significantly improve the system's performance. We evaluated different deep feature extraction algorithms and identified HF-Net, for its robustness, efficiency, and capability to extract both local and global features. Based on these features, we proposed new methods for re-localization and loop closure detection, respectively, both of which have much better performance than the baseline system ORB-SLAM2. The full DXSLAM system gives pose estimates with much lower errors in dynamic environments, even though we do not explicitly address the dynamics. With utilization of the SIMD techniques in modern CPUs, DXSLAM is able to run in real-time on GPU-free mobile computing platforms, enabling easy employment in real robots.

DXSLAM is our first step towards building a \textit{Lifelong SLAM} system and is a proof of the concept of improving SLAM robustness with deep features. There can be two future lines of research. One is to extract better features by improving the CNN architecture and training strategy dedicated for the SLAM use case, such as in \cite{song2020sekd}. The other is to incorporate the deep features into more advanced feature-based SLAM systems, such as ORB-SLAM3 \cite{orb-slam3}. By leveraging the advancement in both deep learning and SLAM, we are going to an era of D$\times$SLAM.

\section*{Acknowledgements}
The authors would like to acknowledge supports from National Natural Science Foundation of China under grant No. 91648116. The authors would also acknowledge support from Beijing Innovation Center for Future Chips, Tsinghua University.

\bibliographystyle{IEEEtran}
\bibliography{references}

\begin{thebibliography}{10}
\providecommand{\url}[1]{#1}
\csname url@samestyle\endcsname
\providecommand{\newblock}{\relax}
\providecommand{\bibinfo}[2]{#2}
\providecommand{\BIBentrySTDinterwordspacing}{\spaceskip=0pt\relax}
\providecommand{\BIBentryALTinterwordstretchfactor}{4}
\providecommand{\BIBentryALTinterwordspacing}{\spaceskip=\fontdimen2\font plus
\BIBentryALTinterwordstretchfactor\fontdimen3\font minus
  \fontdimen4\font\relax}
\providecommand{\BIBforeignlanguage}[2]{{%
\expandafter\ifx\csname l@#1\endcsname\relax
\typeout{** WARNING: IEEEtran.bst: No hyphenation pattern has been}%
\typeout{** loaded for the language `#1'. Using the pattern for}%
\typeout{** the default language instead.}%
\else
\language=\csname l@#1\endcsname
\fi
#2}}
\providecommand{\BIBdecl}{\relax}
\BIBdecl

\bibitem{cadena2016past}
C.~Cadena, L.~Carlone, H.~Carrillo, Y.~Latif, D.~Scaramuzza, J.~Neira, I.~Reid,
  and J.~J. Leonard, ``Past, present, and future of simultaneous localization
  and mapping: Toward the robust-perception age,'' \emph{IEEE Transactions on
  robotics}, vol.~32, no.~6, pp. 1309--1332, 2016.

\bibitem{LoweDistinctive}
D.~G. Lowe, ``Distinctive image features from scale-invariant keypoints,''
  \emph{International Journal of Computer Vision}, vol.~60, no.~2, pp. 91--110.

\bibitem{shi1994good}
J.~Shi \emph{et~al.}, ``Good features to track,'' in \emph{1994 Proceedings of
  IEEE conference on computer vision and pattern recognition}.\hskip 1em plus
  0.5em minus 0.4em\relax IEEE, 1994, pp. 593--600.

\bibitem{rublee2011orb}
E.~Rublee, V.~Rabaud, K.~Konolige, and G.~Bradski, ``{ORB}: An efficient
  alternative to {SIFT} or {SURF},'' in \emph{2011 International conference on
  computer vision}.\hskip 1em plus 0.5em minus 0.4em\relax Ieee, 2011, pp.
  2564--2571.

\bibitem{mur2017orb}
R.~Mur-Artal and J.~D. Tard{\'o}s, ``{ORB-SLAM}2: An open-source {SLAM} system
  for monocular, stereo, and {RGB-D} cameras,'' \emph{IEEE Transactions on
  Robotics}, vol.~33, no.~5, pp. 1255--1262, 2017.

\bibitem{shi2019we}
X.~Shi, D.~Li, P.~Zhao, Q.~Tian, Y.~Tian, Q.~Long, C.~Zhu, J.~Song, F.~Qiao,
  L.~Song \emph{et~al.}, ``Are we ready for service robots? the
  {OpenLORIS-Scene} datasets for lifelong {SLAM},'' in \emph{2020 International
  Conference on Robotics and Automation (ICRA)}, 2020, pp. 3139--3145.

\bibitem{detone2018superpoint}
D.~DeTone, T.~Malisiewicz, and A.~Rabinovich, ``{SuperPoint}: Self-supervised
  interest point detection and description,'' in \emph{Proceedings of the IEEE
  Conference on Computer Vision and Pattern Recognition Workshops}, 2018, pp.
  224--236.

\bibitem{dusmanu2019d2}
M.~Dusmanu, I.~Rocco, T.~Pajdla, M.~Pollefeys, J.~Sivic, A.~Torii, and
  T.~Sattler, ``{D2-Net}: A trainable {CNN} for joint detection and description
  of local features,'' \emph{arXiv preprint arXiv:1905.03561}, 2019.

\bibitem{tang2019gcnv2}
J.~Tang, L.~Ericson, J.~Folkesson, and P.~Jensfelt, ``{GCNv2}: Efficient
  correspondence prediction for real-time {SLAM},'' \emph{IEEE Robotics and
  Automation Letters}, vol.~4, no.~4, pp. 3505--3512, 2019.

\bibitem{yue2019robust}
H.~Yue, J.~Miao, Y.~Yu, W.~Chen, and C.~Wen, ``Robust loop closure detection
  based on bag of {SuperPoints} and graph verification,'' in \emph{2019
  IEEE/RSJ International Conference on Intelligent Robots and Systems
  (IROS)}.\hskip 1em plus 0.5em minus 0.4em\relax IEEE, 2019, pp. 3787--3793.

\bibitem{arandjelovic2016netvlad}
R.~Arandjelovic, P.~Gronat, A.~Torii, T.~Pajdla, and J.~Sivic, ``{NetVLAD}:
  {CNN} architecture for weakly supervised place recognition,'' in
  \emph{Proceedings of the IEEE conference on computer vision and pattern
  recognition}, 2016, pp. 5297--5307.

\bibitem{munoz2020ucoslam}
R.~Mu{\~n}oz-Salinas and R.~Medina-Carnicer, ``{UcoSLAM}: Simultaneous
  localization and mapping by fusion of keypoints and squared planar markers,''
  \emph{Pattern Recognition}, p. 107193, 2020.

\bibitem{huang2019survey}
B.~Huang, J.~Zhao, and J.~Liu, ``A survey of simultaneous localization and
  mapping with an envision in 6g wireless networks,'' 2019.

\bibitem{davison2007monoslam}
A.~J. Davison, I.~D. Reid, N.~D. Molton, and O.~Stasse, ``Monoslam: Real-time
  single camera slam,'' \emph{IEEE transactions on pattern analysis and machine
  intelligence}, vol.~29, no.~6, pp. 1052--1067, 2007.

\bibitem{mur2015orb}
R.~Mur-Artal, J.~M.~M. Montiel, and J.~D. Tardos, ``Orb-slam: a versatile and
  accurate monocular slam system,'' \emph{IEEE transactions on robotics},
  vol.~31, no.~5, pp. 1147--1163, 2015.

\bibitem{qin2018vins}
T.~Qin, P.~Li, and S.~Shen, ``{VINS-Mono}: A robust and versatile monocular
  visual-inertial state estimator,'' \emph{IEEE Transactions on Robotics},
  vol.~34, no.~4, pp. 1004--1020, 2018.

\bibitem{fischer2014descriptor}
P.~Fischer, A.~Dosovitskiy, and T.~Brox, ``Descriptor matching with
  convolutional neural networks: a comparison to sift,'' \emph{arXiv preprint
  arXiv:1405.5769}, 2014.

\bibitem{han2015matchnet}
X.~Han, T.~Leung, Y.~Jia, R.~Sukthankar, and A.~C. Berg, ``{MatchNet}: Unifying
  feature and metric learning for patch-based matching,'' in \emph{Proceedings
  of the IEEE Conference on Computer Vision and Pattern Recognition}, 2015, pp.
  3279--3286.

\bibitem{balntas2016learning}
V.~Balntas, E.~Riba, D.~Ponsa, and K.~Mikolajczyk, ``Learning local feature
  descriptors with triplets and shallow convolutional neural networks.'' in
  \emph{BMVC}, vol.~1, no.~2, 2016, p.~3.

\bibitem{zagoruyko2015learning}
S.~Zagoruyko and N.~Komodakis, ``Learning to compare image patches via
  convolutional neural networks,'' in \emph{Proceedings of the IEEE conference
  on computer vision and pattern recognition}, 2015, pp. 4353--4361.

\bibitem{mishchuk2017working}
A.~Mishchuk, D.~Mishkin, F.~Radenovic, and J.~Matas, ``Working hard to know
  your neighbor's margins: Local descriptor learning loss,'' in \emph{Advances
  in Neural Information Processing Systems}, 2017, pp. 4826--4837.

\bibitem{dai2019comparison}
Z.~Dai, X.~Huang, W.~Chen, L.~He, and H.~Zhang, ``A comparison of {CNN}-based
  and hand-crafted keypoint descriptors,'' in \emph{2019 International
  Conference on Robotics and Automation (ICRA)}.\hskip 1em plus 0.5em minus
  0.4em\relax IEEE, 2019, pp. 2399--2404.

\bibitem{tang2018geometric}
J.~Tang, J.~Folkesson, and P.~Jensfelt, ``Geometric correspondence network for
  camera motion estimation,'' \emph{IEEE Robotics and Automation Letters},
  vol.~3, no.~2, pp. 1010--1017, 2018.

\bibitem{sarlin2019coarse}
P.-E. Sarlin, C.~Cadena, R.~Siegwart, and M.~Dymczyk, ``From coarse to fine:
  Robust hierarchical localization at large scale,'' in \emph{Proceedings of
  the IEEE Conference on Computer Vision and Pattern Recognition}, 2019, pp.
  12\,716--12\,725.

\bibitem{wang2017deepvo}
S.~Wang, R.~Clark, H.~Wen, and N.~Trigoni, ``{DeepVO}: Towards end-to-end
  visual odometry with deep recurrent convolutional neural networks,'' in
  \emph{2017 IEEE International Conference on Robotics and Automation
  (ICRA)}.\hskip 1em plus 0.5em minus 0.4em\relax IEEE, 2017, pp. 2043--2050.

\bibitem{kendall2015posenet}
A.~Kendall, M.~Grimes, and R.~Cipolla, ``Posenet: A convolutional network for
  real-time 6-{DOF} camera relocalization,'' in \emph{Proceedings of the IEEE
  international conference on computer vision}, 2015, pp. 2938--2946.

\bibitem{melekhov2017relative}
I.~Melekhov, J.~Ylioinas, J.~Kannala, and E.~Rahtu, ``Relative camera pose
  estimation using convolutional neural networks,'' in \emph{International
  Conference on Advanced Concepts for Intelligent Vision Systems}.\hskip 1em
  plus 0.5em minus 0.4em\relax Springer, 2017, pp. 675--687.

\bibitem{detone2016deep}
D.~DeTone, T.~Malisiewicz, and A.~Rabinovich, ``Deep image homography
  estimation,'' \emph{arXiv preprint arXiv:1606.03798}, 2016.

\bibitem{angeli2008fast}
A.~Angeli, D.~Filliat, S.~Doncieux, and J.-A. Meyer, ``Fast and incremental
  method for loop-closure detection using bags of visual words,'' \emph{IEEE
  Transactions on Robotics}, vol.~24, no.~5, pp. 1027--1037, 2008.

\bibitem{nister2006scalable}
D.~Nister and H.~Stewenius, ``Scalable recognition with a vocabulary tree,'' in
  \emph{2006 IEEE Computer Society Conference on Computer Vision and Pattern
  Recognition (CVPR'06)}, vol.~2.\hskip 1em plus 0.5em minus 0.4em\relax Ieee,
  2006, pp. 2161--2168.

\bibitem{cummins2008fab}
M.~Cummins and P.~Newman, ``{FAB-MAP}: Probabilistic localization and mapping
  in the space of appearance,'' \emph{The International Journal of Robotics
  Research}, vol.~27, no.~6, pp. 647--665, 2008.

\bibitem{xia2017evaluation}
Y.~Xia, J.~Li, L.~Qi, H.~Yu, and J.~Dong, ``An evaluation of deep learning in
  loop closure detection for visual {SLAM},'' in \emph{2017 IEEE international
  conference on internet of things (iThings) and IEEE green computing and
  communications (GreenCom) and IEEE cyber, physical and social computing
  (CPSCom) and IEEE smart data (SmartData)}.\hskip 1em plus 0.5em minus
  0.4em\relax IEEE, 2017, pp. 85--91.

\bibitem{hou2018bocnf}
Y.~Hou, H.~Zhang, and S.~Zhou, ``{BoCNF}: efficient image matching with bag of
  convnet features for scalable and robust visual place recognition,''
  \emph{Autonomous Robots}, vol.~42, no.~6, pp. 1169--1185, 2018.

\bibitem{kneip2011novel}
L.~Kneip, D.~Scaramuzza, and R.~Siegwart, ``A novel parametrization of the
  perspective-three-point problem for a direct computation of absolute camera
  position and orientation,'' in \emph{CVPR 2011}.\hskip 1em plus 0.5em minus
  0.4em\relax IEEE, 2011, pp. 2969--2976.

\bibitem{fischler1981random}
M.~A. Fischler and R.~C. Bolles, ``Random sample consensus: a paradigm for
  model fitting with applications to image analysis and automated
  cartography,'' \emph{Communications of the ACM}, vol.~24, no.~6, pp.
  381--395, 1981.

\bibitem{sturm2012benchmark}
J.~Sturm, N.~Engelhard, F.~Endres, W.~Burgard, and D.~Cremers, ``A benchmark
  for the evaluation of {RGB-D} {SLAM} systems,'' in \emph{2012 IEEE/RSJ
  International Conference on Intelligent Robots and Systems}.\hskip 1em plus
  0.5em minus 0.4em\relax IEEE, 2012, pp. 573--580.

\bibitem{calonder2010brief}
M.~Calonder, V.~Lepetit, C.~Strecha, and P.~Fua, ``Brief: Binary robust
  independent elementary features,'' in \emph{European conference on computer
  vision}.\hskip 1em plus 0.5em minus 0.4em\relax Springer, 2010, pp. 778--792.

\bibitem{ds-slam}
C.~{Yu}, Z.~{Liu}, X.~{Liu}, F.~{Xie}, Y.~{Yang}, Q.~{Wei}, and Q.~{Fei},
  ``{DS-SLAM}: A semantic visual {SLAM} towards dynamic environments,'' in
  \emph{2018 IEEE/RSJ International Conference on Intelligent Robots and
  Systems (IROS)}, Oct 2018, pp. 1168--1174.

\bibitem{song2020sekd}
Y.~Song, L.~Cai, J.~Li, Y.~Tian, and M.~Li, ``{SEKD}: Self-evolving keypoint
  detection and description,'' \emph{arXiv preprint arXiv:2006.05077}, 2020.

\bibitem{orb-slam3}
C.~Campos, R.~Elvira, J.~J.~G. Rodr{\'\i}guez, J.~M. Montiel, and J.~D.
  Tard{\'o}s, ``{ORB-SLAM3}: An accurate open-source library for visual,
  visual-inertial and multi-map {SLAM},'' \emph{arXiv preprint
  arXiv:2007.11898}, 2020.

\end{thebibliography}

\end{document}